\documentclass[conference]{IEEEtran}
\IEEEoverridecommandlockouts
\usepackage{cite}
\usepackage{stfloats}
\usepackage{tabularx}
\usepackage{amsmath,amssymb,amsfonts}
\usepackage{algorithmic}
\usepackage{graphicx}
\usepackage{textcomp}
\usepackage{amsmath}
\usepackage{array}
\usepackage{multicol}
\usepackage{rotating}
\usepackage{adjustbox}
\usepackage{multirow}
\usepackage{makecell}
\usepackage[table,xcdraw]{xcolor}
\usepackage{booktabs} 
\usepackage{multirow} 
\usepackage{siunitx} 
\def\BibTeX{{\rm B\kern-.05em{\sc i\kern-.025em b}\kern-.08em
    T\kern-.1667em\lower.7ex\hbox{E}\kern-.125emX}}
\begin{document}

\title{Dynamic Sensitivity Filter Pruning using Multi-Agent Reinforcement Learning For DCNN's\\

}

\author{\IEEEauthorblockN{1\textsuperscript{st}Iftekhar Haider Chowdhury}
\IEEEauthorblockA{\textit{Dept. of ECE} \\
\textit{North South University}\\
Dhaka, Bangladesh \\
iftekhar.chowdhury05@northsouth.edu}
\and
\IEEEauthorblockN{2\textsuperscript{nd}Zaed Ikbal Syed}
\IEEEauthorblockA{\textit{Dept. of ECE} \\
\textit{North South University}\\
Dhaka, Bangladesh \\
zaed.ikbal@northsouth.edu}
\and
\IEEEauthorblockN{3\textsuperscript{rd} Ahmed Faizul Haque Dhrubo}
\IEEEauthorblockA{\textit{Dept.of ECE} \\
\textit{North South University}\\
Dhaka, Bangladesh \\
ahmed.dhrubo@northsouth.edu}
\and
\IEEEauthorblockN{4\textsuperscript{th} Mohammad Abdul Qayum}
\IEEEauthorblockA{\textit{Dept.of ECE} \\
\textit{North South University}\\
Dhaka, Bangladesh \\
mohammad.qayum@northsouth.edu}

}

\maketitle

\begin{abstract}
Deep Convolutional Neural Networks (DCNNs) have achieved state-of-the-art performance across various computer vision tasks; however, their practical deployment is limited by computational and memory overhead. This paper introduces Differential Sensitivity Fusion Pruning (DSFP), a novel single-shot filter pruning framework that focuses on evaluating the stability and redundancy of filter importance scores across multiple criteria. DSFP computes a differential sensitivity score for each filter by fusing the discrepancies among gradient-based sensitivity, first-order Taylor expansion, and KL-divergence of activation distributions. An exponential scaling mechanism is applied to emphasize filters with inconsistent importance across metrics — identifying candidates that are structurally unstable or less critical to the model’s performance. Unlike iterative or reinforcement learning-based pruning strategies, DSFP is efficient and deterministic, requiring only a single forward-backward pass for scoring and pruning. Extensive experiments across varying pruning rates (50\%–70\%) demonstrate that DSFP significantly reduces model complexity—achieving over 80\% FLOPs reduction—while maintaining high accuracy. For instance, at 70\% pruning, our approach retains up to 98.23\% of baseline accuracy, surpassing traditional heuristics in both compression and generalization. The proposed method presents an effective solution for scalable and adaptive DCNN compression, paving the way for efficient deployment on edge and mobile platforms.
\end{abstract}


\section{Introduction}
Deep Convolutional Neural Networks (DCNNs) have become foundational to modern computer vision systems, powering applications in image classification, object detection, and semantic segmentation.\cite{b1,b2,b3} with foundational works like AlexNet\cite{b4} and VGG revolutionizing the field. While these models demonstrate remarkable performance, their increasing depth and complexity lead to excessive computational cost, memory usage, and energy consumption. Such requirements pose significant challenges for deploying DCNNs on edge devices and in real-time applications where resources are limited. Model compression techniques such as pruning, quantization, and knowledge distillation have emerged as viable solutions to this challenge. Among these, filter pruning has garnered particular attention due to its structural compatibility with hardware acceleration and ability to directly reduce computational overhead. Traditional filter pruning strategies rely on hand-crafted sensitivity metrics or heuristics, such as weight magnitudes or Taylor-based approximations, to estimate the contribution of each filter to the overall network performance. However, these methods are typically static, fail to generalize across network layers, and often ignore the evolving dynamics of training.

To address these limitations, this paper proposes a novel approach: Dynamic Sensitivity Filter Pruning (DSFP). Instead of relying on static heuristics, DSFP evaluates filter importance post-training using a fused sensitivity metric that combines gradient-based scores, first-order Taylor approximations, and Kullback-Leibler (KL) divergence. A central innovation is the use of an exponential fusion function, which accentuates disparities among the metrics to better highlight critical filters. This enables more informed and aggressive pruning decisions while preserving model accuracy. By decoupling pruning from training and introducing dynamic, data-driven importance estimation, DSFP offers a stable and adaptable framework for efficient model compression. We validate the effectiveness of DSFP through extensive experiments on standard benchmarks and pruning scenarios. Our results demonstrate that the proposed method achieves substantial reductions in floating-point operations (FLOPs) and parameter counts, while preserving, and in some cases enhancing, model accuracy post-pruning. Notably, our approach achieves up to 96.92\% of original accuracy at 70\% pruning, outperforming conventional pruning baselines.

In summary, the key contributions of this work are:
\begin{enumerate}
    \item We propose \textbf{Differential Sensitivity Filter Pruning (DSFP)}, a single-shot pruning strategy designed for efficient model compression using dynamic, layer-wise filter evaluation.
    \item We design a novel \textbf{exponential sensitivity fusion function} that robustly combines gradient sensitivity, Taylor expansion, and KL-divergence to guide pruning decisions.
    \item We empirically validate DSFP on standard benchmarks, showing superior performance compared to traditional filter pruning methods in terms of both accuracy retention and FLOPs reduction.
\end{enumerate}

\section{Related Works}
Model compression has emerged as a critical area of research to address the growing computational demands of deep learning models. Among various techniques, filter pruning stands out for its ability to remove redundant filters in a structured manner, thereby reducing model size without necessitating specialized hardware or software for deployment.

Early approaches in filter pruning leveraged simple heuristics such as the $\ell_1$-norm of filter weights\cite{b5,b6,b7}, under the assumption that filters with smaller weights contribute less to overall network performance\cite{b8}. Molchanov et al.\ introduced a more principled method based on first-order Taylor expansion, estimating the change in loss incurred by pruning individual filters. Similarly, gradient-based importance metrics were proposed to capture the average absolute contribution of filters during training\cite{b9}. Comprehensive surveys\cite{b10,b11} have since cataloged these and other techniques. While effective in practice, many of these methods treat pruning as a static, one-size-fits-all operation, often failing to generalize across heterogeneous layers or adapt to the underlying training dynamics. Heuristic metrics are typically scale-sensitive and layer-agnostic, leading to suboptimal pruning in deep convolutional architectures.

To address these limitations, we propose \textbf{Differential Sensitivity Filter Pruning (DSFP)}, a single-shot, data-driven framework for structured filter pruning. Our method introduces a novel fused importance score that integrates gradient sensitivity, Taylor expansion, and KL-divergence between activation distributions using an exponential fusion function. This fused score enhances robustness across layers and magnifies meaningful discrepancies in filter relevance, resulting in more informed pruning decisions. Additionally, we incorporate a lightweight, bandit-style pruning rate tuner that adaptively adjusts the pruning ratio for each layer. This scalar learner updates its pruning preferences based on empirical rewards from post-pruning fine-tuning, striking a balance between compression and accuracy without relying on complex reinforcement learning infrastructure. Together, the fused evaluation and bandit-guided tuning enable DSFP to efficiently prune filters in a single pass, while retaining strong generalization across networks and datasets.

\section{Methodology}

The proposed approach introduces a dynamic filter pruning framework called Differential Sensitivity Fusion Pruning (DSFP), where the pruning process is modeled as a Markov Decision Process (MDP). The method operates in a post-training setting and is composed of three key components: (A) filter importance estimation using a composite sensitivity fusion metric, (B) layer-wise pruning decision-making using an MDP-based agent that selects pruning ratios based on learned Q-values and exploration strategies, and (C) fine-tuning the pruned model with knowledge distillation to recover potential accuracy loss.

\subsection{Filter Importance Estimation}

To quantify the importance of individual filters, we employ a composite metric based on three sensitivity analysis techniques:

\begin{enumerate}
    \item \textbf{Mean Absolute Gradient (Grad(F))}: Measures the average magnitude of gradients with respect to each filter, indicating its contribution to the learning dynamics.
    \[
    \text{Grad}(F) = \frac{1}{N} \sum_{i=1}^{N} \left| \frac{\partial L}{\partial F_i} \right|
    \]

    \item \textbf{Taylor Expansion (Taylor(F))}: Estimates the impact of filter removal on loss using a first-order Taylor approximation.
    \[
    \text{Taylor}(F) = \sum_{i=1}^{N} \left| \frac{\partial L}{\partial F_i} \cdot F_i \right|
    \]

    \item \textbf{KL-Divergence (KL(F))}: Captures the divergence between the original model’s output distribution and the pruned version.
    \[
    \text{KL}(F) = D_{KL}(p(F) \parallel q(F))
    \]
\end{enumerate}

These metrics are fused into a unified importance score using a differential sensitivity fusion function:
\[
\text{Imp}(F) = e^{|\text{Grad}(F) - \text{Taylor}(F)|} + e^{|\text{Taylor}(F) - \text{KL}(F)|} + \frac{1}{2} e^{|\text{Grad}(F) - \text{KL}(F)|}
\]
This formulation captures the pairwise discrepancies between three complementary importance metrics—gradient sensitivity, first-order Taylor approximation, and KL-divergence. Filters with inconsistent scores across these criteria are emphasized, guiding the pruning process to retain structurally or semantically critical components.

\subsection{Adaptive Pruning Ratio Selection via Scalar Reward Regression}

To determine optimal layer-wise pruning ratios without the overhead of full reinforcement learning or joint multi-agent coordination, we adopt a lightweight, bandit-inspired scalar regression approach. This method treats pruning ratio selection as a one-shot decision problem, where the goal is to learn a predictive mapping from a global context to the optimal compression level per layer.

\begin{itemize}
    \item \textbf{State:} A fixed scalar context (e.g., $s = 50$) encodes a desired global pruning intensity. This abstraction simplifies the input space and avoids modeling temporal dynamics.

    \item \textbf{Action:} For each convolutional layer, the output action is a continuous pruning ratio $a \in [10, 90]$. An $\varepsilon$-greedy exploration strategy perturbs the ratio within a small window during training to allow for policy refinement.

    \item \textbf{Reward:} After applying the proposed pruning ratios and optionally fine-tuning, a scalar reward is computed to capture the trade-off between preserved accuracy and computational savings.

    \item \textbf{Learning:} A simple regression model is trained using collected $(s, a, r)$ tuples to predict expected reward values. By minimizing the mean squared error between predicted and actual rewards, the model gradually improves its ability to suggest effective pruning ratios.
\end{itemize}

This approach resembles contextual bandit optimization and forgoes sequential value updates, policy learning, or multi-agent coordination\cite{b12}. It offers an efficient and scalable mechanism for globally adapting compression levels across network layers with minimal computational cost. While conceptually simpler than reinforcement learning, this technique complements the filter-level MARL pruning\cite{b13} by guiding the global compression policy in a data-driven manner.

\subsection{Knowledge Distillation}

Unlike iterative pruning during training, our approach follows a \textbf{post-training pruning} strategy, which offers more stability in evaluating filter importance and avoids disrupting the learning dynamics. The overall procedure consists of the following steps:

\begin{enumerate}
    \item \textbf{Pretraining}: The DCNN is fully trained to convergence on the target task to learn optimal feature representations.
    \item \textbf{Sensitivity Analysis}: After training, the importance scores of all filters are computed using the composite sensitivity function described in Section III-A.
    \item \textbf{Pruning via Sensitivity-Guided Q-Learning}: Layer-wise pruning decisions are made using a lightweight agent-based Q-learning mechanism, where each agent selects pruning ratios based on filter importance scores derived from the Differential Sensitivity Fusion metric, similar to the strategy used in QLP\cite{b14}. Agents balance exploration and exploitation to dynamically determine optimal compression rates for their respective layers.
    \item \textbf{Fine-Tuning with Knowledge Distillation}: To recover potential accuracy loss after pruning, the pruned model is fine-tuned using knowledge distillation\cite{b15}. In this process, the original pretrained model (teacher) provides soft targets to guide the pruned model (student), enabling it to mimic the richer output distributions and retain better generalization.
\end{enumerate}

By integrating knowledge distillation into the fine-tuning phase, the pruned model achieves higher post-pruning accuracy compared to conventional fine-tuning methods. This combination ensures that pruning is both aggressive and performance-preserving, making the resulting network more efficient without significant degradation in predictive capability.

This delayed pruning strategy allows for more stable evaluation of filter contributions and avoids interference with the network’s training dynamics.

\begin{figure*}[t]
\centering
\includegraphics[width=0.7\textwidth]{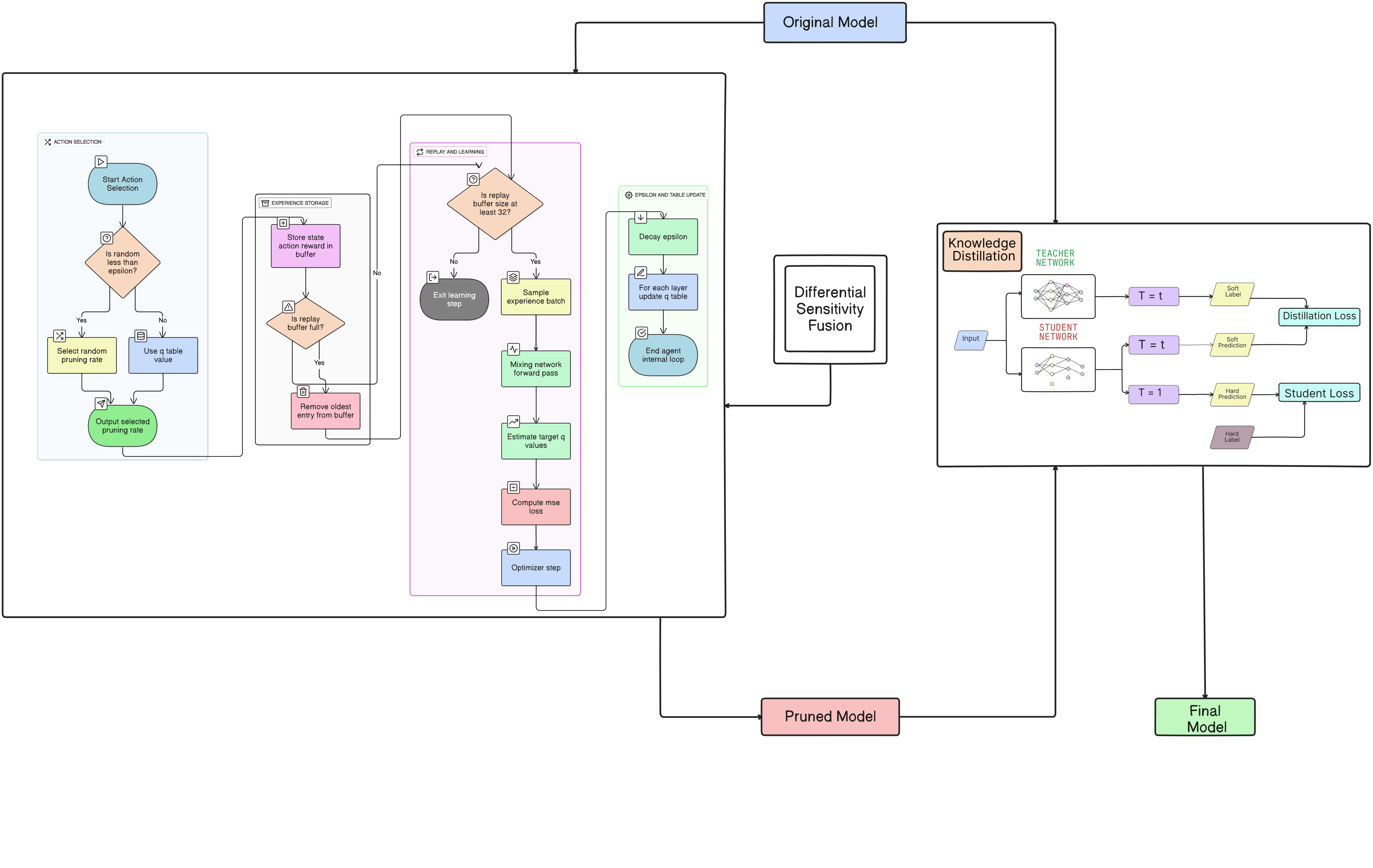}
\caption{Architecture of the proposed system design.}
\label{fig:jv}
\end{figure*}

In figure 1 the process begins with sensitivity analysis using a differential sensitivity fusion function that combines gradient, Taylor expansion, and KL divergence metrics to compute filter importance. A Q-learning–inspired pruning controller explores candidate pruning rates using an $\varepsilon$-greedy strategy and updates a Q-table via experience replay and loss minimization. This guides layer-wise pruning decisions in a post-training setting. Finally, knowledge distillation is employed to fine-tune the pruned model by transferring soft and hard label information from the original (teacher) network to the student network, resulting in the final compact model.

\section{Experiment}
\subsection{Setup}

This section details the experimental setup used to evaluate the DSFP framework on structured filter pruning tasks.

\textbf{Dataset:} 
We perform experiments on the CIFAR-10 dataset\cite{b16}, which contains 60,000 color images divided into 10 classes, with 50,000 images for training and 10,000 images for testing. Each image has a resolution of $32\times32$ pixels. CIFAR-10 serves as a standard benchmark for evaluating network pruning techniques due to its moderate complexity and manageable size.

\textbf{Baseline Architectures:} 
Our experiments utilize two widely recognized deep learning models: \textbf{AlexNet}\cite{b4} and \textbf{VGG16}\cite{b17}, both adapted for CIFAR-10. AlexNet represents a moderately deep architecture suitable for exploring early-stage pruning behavior, whereas VGG16\cite{b18} provides a deeper and more challenging structure for evaluating aggressive pruning under higher compression rates.

\textbf{Implementation Details:} 
The models are fine-tuned using stochastic gradient descent (SGD) with momentum 0.9 and a weight decay of $5\times10^{-4}$. For learning rate scheduling, Cosine Annealing with Warm Restarts is employed, with an initial warm-up period of $T_0=50$ epochs and a doubling cycle multiplier ($T_{mult}=2$). The learning rate is initialized at 0.001 for standard fine-tuning and adjusted dynamically by the scheduler during training. 

Gradient accumulation over four mini-batches is used to simulate larger batch sizes, allowing stable updates without exceeding GPU memory limitations. Mixed precision training is utilized via automatic mixed precision (AMP) and gradient scaling to accelerate training and reduce memory consumption. 

To further enhance the base model before pruning, mixup data augmentation is applied during initial training, mixing pairs of images and corresponding labels to improve model robustness.
\par \textbf{Pruning Settings:} 
Post-training pruning is performed by first computing sensitivity scores using the proposed Differential Sensitivity Fusion metric, which combines Gradient, Taylor expansion, and KL-divergence metrics through an exponential fusion strategy, as described in Section III-A. Based on these scores, lightweight Q-learning agents make pruning decisions independently for each layer, selecting optimal compression rates. We evaluate pruning performance at compression rates of 50\%, 60\%, and 70\% of the filters.

\textbf{Fine-Tuning and Knowledge Distillation:} 
After pruning, the student model (pruned network) is fine-tuned under the supervision of the original fully trained model (teacher) using an improved knowledge distillation (KD) strategy. 

The KD loss combines a softened Kullback-Leibler divergence between the student and teacher output distributions and a standard cross-entropy loss with label smoothing. A dynamic balancing coefficient $\alpha$ is employed, which linearly decreases over the course of training to gradually shift focus from distillation supervision to direct supervision. The temperature parameter for softening the teacher logits is set to 4.0.

The optimizer used for KD fine-tuning is AdamW with an initial learning rate of $1\times10^{-4}$, combined with a Cosine Annealing Warm Restarts scheduler ($T_0=10$, $T_{mult}=2$). The best-performing pruned student model is saved based on its validation accuracy throughout training.

\textbf{Evaluation Metrics:}
The performance of each model is evaluated based on three criteria: (i) Top-1 classification accuracy on the CIFAR-10 test set, (ii) number of remaining convolutional filters, and (iii) total number of learnable parameters (in millions). These metrics provide a comprehensive view of the compression-accuracy trade-off achieved by the proposed pruning method.

\subsection{Results and Analysis}

\begin{table}[htbp]
\centering
\caption{Pruning results of Alexnet on CIFAR-10}
\label{tab:performance_comparison}
\scriptsize 
\setlength{\tabcolsep}{3pt} 
\begin{tabular}{|c|c|c|c|c|}
\hline
Method & Acc. (pruned) & Acc (finetuned) & Filters & Params (M) \\ \hline
Base & 87.76 & - & 1152 & 6.9768 \\ 
Sensitivity    (50\%) & 44.28 & 87.40 (99.58\%) & 585 & 5.8021 \\ 
Sensitivity    (60\%) & 33.11 & 86.66 (98.75\%) & 443 & 5.6191 \\ 
Sensitivity    (70\%) & 33.71 & 86.21 (98.23\%) & 317 & 5.3391 \\ \hline
\end{tabular}
\end{table}

\begin{table}[htbp]
\centering
\caption{Pruning results of VGG16 on CIFAR10}
\label{tab:performance_comparison}
\scriptsize 
\setlength{\tabcolsep}{3pt} 
\begin{tabular}{|c|c|c|c|c|}
\hline
Method & Acc. (pruned) & Acc (finetuned) & Filters & Params (M) \\
\hline
Base & 93.76 & - & 4224 & 14.8486 \\
Sensitivity (50\%) & 11.91 & 93.07 (99.26\%) & 2154 & 7.5945 \\
Sensitivity (60\%) & 10.04 & 92.76 (98.93\%) & 1754 & 6.1164 \\
Sensitivity (70\%) & 10.00 & 91.45 (97.53\%) & 1339 & 4.7087 \\
\hline
\end{tabular}
\end{table}

We initially conducted experimental analysis on the pruning performance of VGG-16 using the benchmark CIFAR-10 dataset. To further validate the robustness and generalizability of the proposed Differential Sensitivity Fusion importance method, we subsequently extended our experiments by replacing the VGG-16 architecture with AlexNet. 

VGG-16 + CIFAR-10:
\par In this work, we investigate the pruning performance of VGG-16 on CIFAR-10 using a customized Differential Sensitivity Fusion importance method. To assess model compression under varying sparsity levels, we target pruning rates of 50\%, 60\%, and 70\%. Unlike traditional approaches relying on multiple importance metrics, we exclusively employ our proposed fusion of gradient sensitivity, Taylor expansion, and KL divergence to determine filter importance.

The experimental results are presented in Table 2. The "Base" row reports the metrics of the original, fully fine-tuned VGG-16 model. The "Acc. (pruned)" column reflects the classification accuracy immediately after pruning, while the "Acc. (fine-tuned)" column shows the final accuracy after fine-tuning with Knowledge Distillation, where the original model serves as the teacher. Following each pruning stage, the model is fine-tuned for 300 epochs to ensure performance recovery. The percentage in parentheses indicates the proportion of original accuracy retained after fine-tuning.

Our results demonstrate that at a 50\% pruning rate, the fine-tuned network recovers 99.26\% of the original accuracy, achieving 93.07\%. As the pruning rate increases to 60\%, the fine-tuned model maintains 98.93\% of the baseline accuracy, and at 70\% pruning, 97.53\% is preserved. These findings highlight that even under substantial pruning, the network retains strong predictive performance.

In addition to accuracy preservation, significant reductions in model size and computational cost are observed. For example, at 50\% pruning, the number of parameters decreases from 14.85M to 7.59M, with FLOPs reduced from 0.0047M to 0.0028M. As the pruning rate increases, both parameter count and FLOPs continue to decline, underscoring the efficacy of our approach in achieving compact models without substantial performance loss.

\begin{figure}[htbp]
  \centering
  \includegraphics[width=1\columnwidth]{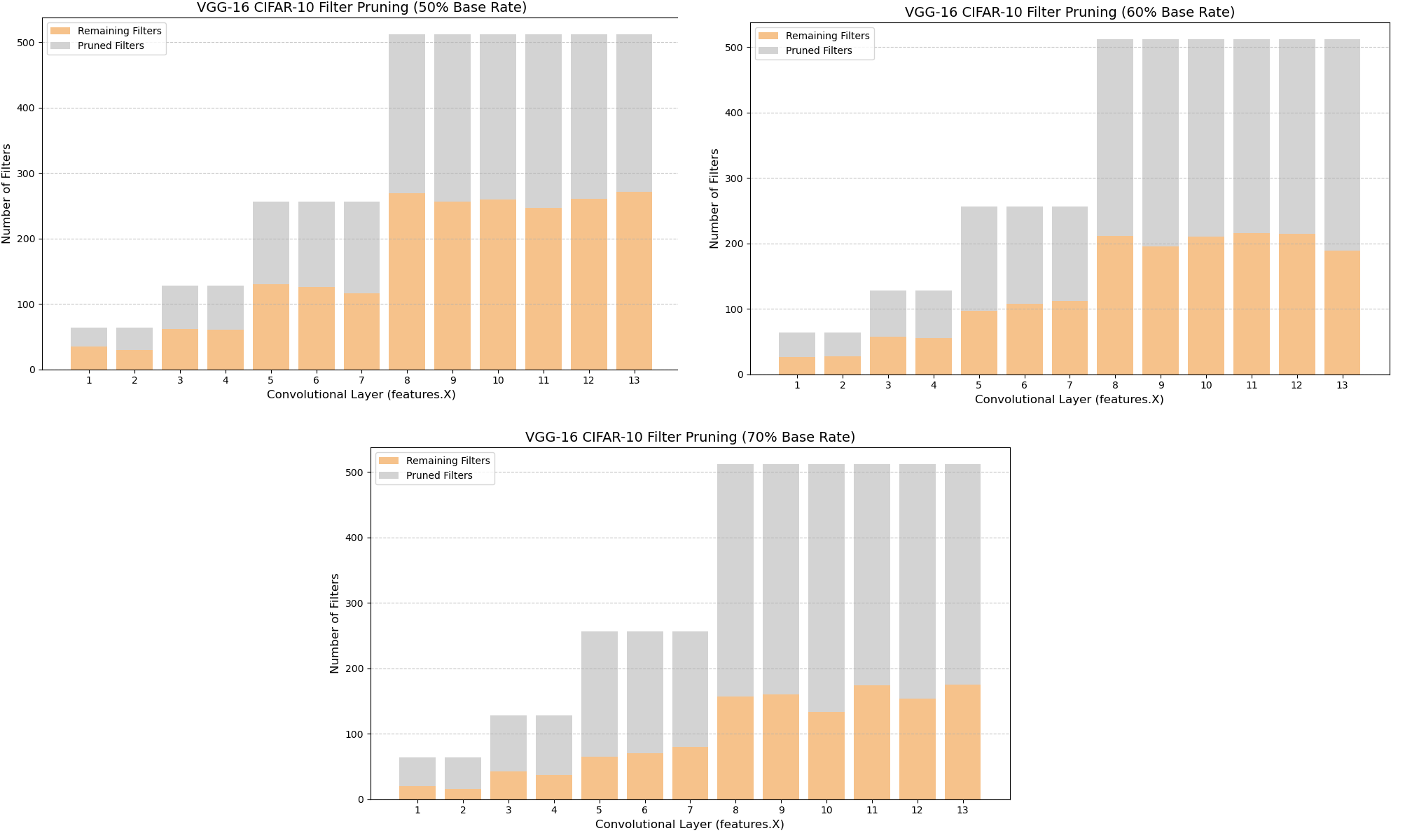} 
  \caption{Filter pruning distribution across convolutional layers in VGG-16 on CIFAR-10 at different base pruning rates.}
  \label{fig:j}
\end{figure}

In Figure 2 (a) At a 50\% base pruning rate, pruning is evenly spread across layers, with deeper layers (8–13) still retaining a significant number of filters (around 250–270). (b) At a 60\% base rate, the number of remaining filters further decreases, especially in early layers like 1 and 2, while deeper layers retain around 210–220 filters. (c) At a 70\% base rate, aggressive pruning is evident throughout, with early layers like 1–3 retaining very few filters (30–60), and even the deeper layers reduced to approximately 150–180 filters.

AlexNet + CIFAR-10:
\par Following our evaluation on VGG-16, we further assess the performance of our pruning strategy on the AlexNet architecture, again using the CIFAR-10 dataset. Consistent with the previous setup, we apply pruning rates of 50\%, 60\%, and 70\% based on the proposed sensitivity-driven importance method. As before, filter importance is determined without relying on multiple separate metrics, instead using sensitivity analysis to guide pruning decisions.

The results are summarized in Table 1. The "Base" model refers to the fully trained AlexNet without any pruning. The "Acc. (pruned)" column indicates the classification accuracy immediately after pruning, while "Acc. (fine-tuned)" shows the accuracy after applying fine-tuning with Knowledge Distillation, where the original unpruned network serves as the teacher. For each pruning level, the pruned model was initially fine-tuned for 300 epochs.

Even at 70\% pruning rate, the model still retained an accuracy of 98.24\% of the original model, the number of parameters decreases from 6.98M to 5.8M, with FLOPs reduced from  0.0014M to 0.0009M. At 50\% pruning rate, the pruned model does exceptionally well by retaining 99.58\% of the base model's accuracy.

\begin{figure}[htbp]
  \centering
  \includegraphics[width=1\columnwidth]{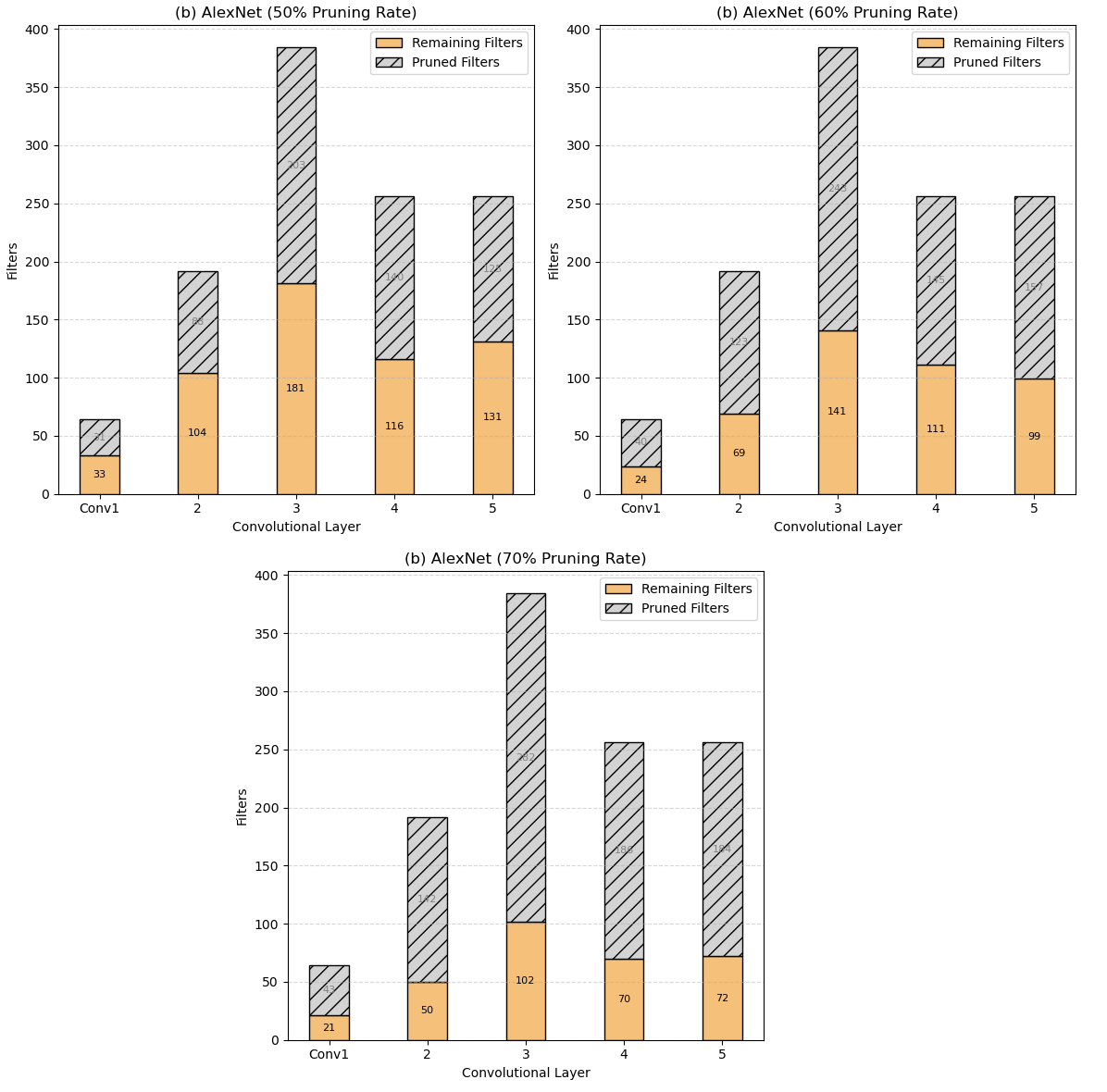}
  \caption{Filter pruning distribution across convolutional layers in AlexNet at different pruning rates.}
  \label{fig:j}
\end{figure}

In Figure 3 (a) At the pruning rate 50\%, layer 3 retains the highest number of filters (181), while Conv1 retains the fewest (33). (b) At 60\% pruning rate, layer 3 still retains the most filters (141), while Conv1 decreases to 24 and layer 2 to 69. (c) At 70\% pruning rate, the most aggressive pruning is observed across all layers, with layer 3 still maintaining the highest number of remaining filters (102). Conv1 retains only 21 filters, while layers 4 and 5 preserve 70 and 72 filters respectively.

\subsection{Discussion}
 
\begin{figure}[htbp]
  \centering
  \includegraphics[width=1\columnwidth]{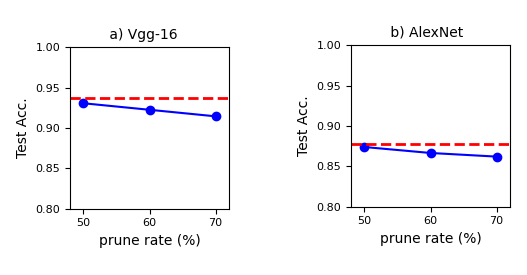}
  \caption{Effect of pruning on test accuracy for (a) VGG-16 and (b) AlexNet.}
  \label{fig:j}
\end{figure}

In Figure 4 the red dashed lines indicate the baseline (unpruned) test accuracy. As the pruning rate increases from 50\% to 70\%, both models show a decrease in the precision of the test. AlexNet experiences a mild drop, while VGG-16 exhibits a more noticeable degradation, indicating its greater sensitivity to pruning. The experimental results on both VGG-16 and AlexNet demonstrate the effectiveness of the proposed Differential Sensitivity Fusion importance method for structured filter pruning. Across different architectures and pruning rates, the method consistently achieves significant reductions in model complexity while preserving a high percentage of the original classification accuracy.

In the case of VGG-16, the proposed method was able to retain over 97\% of the baseline accuracy even at a 70\% pruning rate, demonstrating its ability to accurately identify and preserve the most critical filters. Similarly, for AlexNet, despite the network's different architecture and parameter distribution, the method achieved comparable levels of performance retention after pruning and extended fine-tuning. Notably, the necessity of longer fine-tuning (up to 700 epochs) for AlexNet at higher pruning rates highlights the interplay between model structure, pruning severity, and the optimization dynamics during recovery. The success of the method across architectures with distinct designs—such as the deeper, narrower VGG-16 and the shallower, wider AlexNet—indicates that the fusion of gradient sensitivity, Taylor expansion, and KL-divergence provides a robust, architecture-agnostic measure of filter importance. Unlike traditional pruning strategies that often rely on isolated importance criteria, our integrated approach captures a broader range of information regarding a filter’s contribution to the network's predictive capabilities.

In future work, it would be valuable to further evaluate the generalization ability of the proposed Differential Sensitivity Fusion method by applying it to a broader range of convolutional neural network architectures, such as VGG-19, ResNet, and other modern deep learning models.

\section{Conclusion}
In this study, we introduced a novel filter importance evaluation technique, termed Differential Sensitivity Fusion, which integrates gradient sensitivity, Taylor expansion, and KL-divergence to guide the structured pruning of deep convolutional neural networks. Experimental validations on VGG-16 and AlexNet using the CIFAR-10 dataset demonstrated that the proposed method achieves substantial model compression while maintaining minimal loss in classification accuracy.

Extensive experiments across varying pruning rates highlight the effectiveness of our method in identifying and preserving critical filters. Pruned networks were able to retain over 97\% of their original performance, even under aggressive pruning scenarios. To further mitigate performance degradation after pruning, Knowledge Distillation (KD) was employed during fine-tuning, allowing the pruned models to recover more effectively. The successful application of the proposed method across architectures of differing complexity and depth underscores its robustness and generalizability.

Future work will focus on extending the evaluation of the Differential Sensitivity Fusion method to more complex architectures, such as VGG-19 and the ResNet family\cite{b1}, as well as exploring its applicability to modern Transformer-based models like ViT\cite{b19}, and investigating its effectiveness on vision tasks beyond image classification. Additionally, further research into enhancing KD-based fine-tuning strategies could lead to even more efficient and resilient model deployment in real-world applications.

\vspace{12pt}
\color{red}

\end{document}